\documentclass[conference]{IEEEtran}
\IEEEoverridecommandlockouts
\usepackage{cite}
\usepackage{amsmath,amssymb,amsfonts}
\usepackage{algorithmic}
\usepackage{graphicx}
\usepackage{textcomp}
\usepackage{xcolor}
\ifCLASSOPTIONcompsoc
\usepackage[nocompress]{cite}
\else
\usepackage{cite}
\fi

\usepackage{times} 
\usepackage{helvet} 
\usepackage{courier} 
\usepackage{caption}
\usepackage{algorithm}
\usepackage[utf8]{inputenc}
\usepackage{newfloat}
\usepackage{listings}
\usepackage{verbatim}
\usepackage{latexsym}
\usepackage{booktabs}
\usepackage[english]{babel}
\usepackage{xspace}
\usepackage{psfrag}
\usepackage[misc]{ifsym}
\usepackage{environ}

\newenvironment{nalign}{
\begin{equation}
\begin{aligned}[b]
 }{
     \end{aligned}
     \end{equation}
     \ignorespacesafterend
 }

\newcommand{\equ}{\begin{nalign}}
\newcommand{\eque}{\end{nalign}}

\newcommand{\N}{\ensuremath{\mathbb{N}}\xspace}

\newtheorem{Definition}{Definition}[section]
\newtheorem{Lemma}{Lemma}[section]
\newtheorem{Theorem}{Theorem}[section]
\newtheorem{Example}{Example}[section]
\newtheorem{Note}{Remark}[section]

\newcommand{\bea}{\begin{eqnarray*}}
\newcommand{\ea}{\end{eqnarray*}}
\newcommand{\be}{\begin{equation}}
\newcommand{\ee}{\end{equation}}

\newcommand{\Def}{\begin{Definition}}
\newcommand{\Defa}{\end{Definition}}
\newcommand{\Lem}{\begin{Lemma}}
\newcommand{\Lema}{\end{Lemma}}
\newcommand{\Thea}{\begin{Theorem}}
\newcommand{\Thee}{\end{Theorem}}
\newcommand{\Ex}{\begin{Example}}
\newcommand{\Exa}{\end{Example}}
\newcommand{\Nb}{\begin{Note}}
\newcommand{\Ne}{\end{Note}}

\begin{document}

\title{Outline of an Independent Systematic Blackbox Test for ML-based Systems\\
\thanks{This work has been funded by https://ki-lok.itpower.de/ (German Federal Ministry for Economic Affairs and Climate Action and managed by TÜV Rheinland (project no.: 19121007A)) and https://iml4e.org (German Federal Ministry for Education and Research (project no.: 01IS21021C))}
}

\author{\IEEEauthorblockN{1\textsuperscript{st} Hans-Werner Wiesbrock}
\IEEEauthorblockA{\textit{ITPower Solutions GmbH} \\
Kolonnenstrasse 26\\
10829 Berlin, Germany \\
hans-werner.wiesbrock@itpower.de}
\and
\IEEEauthorblockN{2\textsuperscript{nd} Jürgen Großmann}
\IEEEauthorblockA{\textit{Fraunhofer-Institut für Offene Kommunikationssysteme FOKUS} \\
Kaiserin-Augusta-Allee 31,\\
10589 Berlin, Germany \\
juergen.grossmann@fokus.fraunhofer.de}
}

\maketitle

\begin{abstract}
This article proposes a test procedure that can be used to test ML models and ML-based systems independently of the actual training process. 
In this way, the typical quality statements such as accuracy and precision of these models and system can be verified independently, taking into account their black box character and the immanent stochastic properties of ML models and their training data. 
The article presents first results from a set of test experiments and suggest extensions to existing test methods reflecting the stochastic nature of ML models and ML-based systems.
\end{abstract}

\begin{IEEEkeywords}
	Testing AI Systems, Blackbox Test for AI Systems, Systematic Evaluation of Training Datasets, Probabilistic Modelling
\end{IEEEkeywords}
\section{Introduction}\label{section:Introduction}

Machine Learning (ML) and ML-based systems are used today in a wide range of areas, and increasingly also in safety-critical domains.
Their range of application is growing exponentially. At the same time, more and more experts are warning of the uncertainties and risks associated with the uncontrolled and overly rapid development of AI systems \cite{OpenLetter}.
In general, there is a growing need to provide methods and procedures for testing functioning and quality characteristics of these systems.

Various methods currently exist to test and verify ML-based systems, be it formal verification, simulation approaches or classical testing \cite{Albarghouthi.21.09.2021,Jackson.29.03.2021,VasuSingh.}, or new analysis methods in the context of XAI \cite{Hoyer.30.07.2019,Guidotti.06.02.2018b}. The methods aim for providing evidence on the robustness and trustworthiness of the ML models or ML-based system. 
However, the complexity and variety of possible inputs limit the applicability of formal methods and XAI to simple situations and samples. Furthermore, in the case of simulations, neither end-of-test criteria nor the required test case distributions are systematically specified.

Similar to the traditional development of complex software systems, testing has proven to be the most effective method for proving quality and gaining trust in ML. While there already exist test approaches for ML-based systems, see for example \cite{Marselis.b,Zhang.20.06.2019b}, we have not yet recognized a training process independent, systematic test approach that is also capable of verifying typical quality characteristics such as accuracy and precision in a way that respects the stochastic nature of a data-based inference process.

In the traditional system development process, it is of paramount importance that tests are specified independently of the design and development activities. The agreed end-of-test criteria indicate the thoroughness with which the system under test (SUT) is tested and determine when the test process can be completed. However, a comparable test process that distinguishes between training and testing does not yet exist for ML-based systems.

This paper presents a novel, comprehensive methodology for testing and evaluating the performance of ML-based black box systems, addressing some of the limitations of existing approaches.
In order to demonstrate the feasibility of the proposed approach, a preliminary experiment was conducted. The experiment involved utilising an ML-based object detection system, based on the Coco dataset \cite{linMicrosoftCOCO2015}, with the centernet hourglass model, see \cite{duanCenterNetKeypoint2019}, serving as the system under investigation.

However, our proposed approach can easily be transferred to other systems gained through supervised and unsupervised training. Reinforcement Learning (RL) systems and Generative Systems however, require separate consideration and are reserved for later work.

\textit{This publication makes the following contributions:}
\begin{itemize}
  \item We introduce the novel concept of \textit{probabilistically extended ontologies}, which enables the independent testing of standard statistical quality criteria for ML-based systems, such as their accuracy and precision, irrespective of their development or complexity.
\item This concept allows us to formulate the \textit{uniformity hypothesis}, which serves as a final criterion for the refinement of ontologies.
  \item We obtain a probability model for the Operational Design Domain (ODD) of a ML system, from which we can derive end-of-test criteria depending on the required quality of the system.
  \item We integrate these concepts into a testing pipeline that can be performed independently of the model development and training using the black-box method.
\end{itemize}
If data scientists, developers and testers have worked well and independently, this process will be able to reproduce the statistical quality criteria provided in the training process and required by the user.

\Nb \label{note:TraingData}
The development of supervised learning requires a large amount of training data.
It should be \textit{complete} and \textit{representative} with respect to the Operational Design Domain (ODD), a non-trivial undertaking.
The concept of \textit{probabilistically extended ontologies} can assist developers and data scientists, as it makes the various assumptions and implicit guiding ideas of the  data selection process explicit.
The ontology helps to achieve completeness, the probabilistic extension to achieve representativeness.
It should be emphasised, however, that this work is not concerned with the specification and tagging of training or test data as required in development and training, but for an independent test.
\Ne

\subsection{Related work}
\label{section:relatedWork}
Traditional software testing has major limitations in terms of the dynamics of machine learning, the sheer size of the problem domain, the underlying oracle problem, and the defintion of completenes and realiable test end criteria \cite{weyuker.oracle}. The oracle problem is currently often addressed by metamorphic testing, see Murphy et. al \cite{murphy_improving_2008}, Segura \cite{segura_survey_2016} and, Xie \cite{xie_testing_2011}. 

To define test end criteria and completeness, there are a number of proposals that combine systematic testing of a DNN component with coverage criteria related to the structure of DNNs. These include simple neuron coverage by Pei et al. \cite{pei_deepxplore:_2017}, which considers the activation of individual neurons in a network as a variant of statement coverage. Ma et al. \cite{ma_deepgauge_2018} define additional coverage criteria that follow a similar logic to neuron coverage and focus on the relative strength of the activation of a neuron in its neighbourhood. Motivated by the MC/DC tests for traditional software, Sun et al. \cite{sun_structural_2019} proposes an MC/DC variant for DNNs, which establishes a causal relationship between neurons clustering i.e., the features in DNNs. The core idea is to ensure that not only the presence of a feature, but also the combination of complex features from simple feature needs to be tested. Wicker et al \cite{wicker_feature-guided_2018} and Cheng et al. \cite{cheng_towards_2018} refer to partitions of the input space as coverage items, so that coverage measures are defined considering essential properties of the input data distribution. While Wicker et al. discretizes the input data space into a set of hyper-rectangles, in Cheng al. it is assumed that the input data space can be partitioned along a set of weighted criteria to describe the operating conditions.

Often ontologies have been successfully used to model the Operational Design Domain (ODD) and thus the input domain in the context of autonomous agents such as automated vehicles and automated trains \cite{huang_ontology-based_2019, armand_ontology-based_2014}. 
These ontologies represent the relevant contextual information about an agents operation, such as the environmental conditions, geographical features, traffic rules, and operational constraints. This can help in defining the boundaries and limitations of the automated agents operation, ensure that the agent stays within its operational design domain and provide a good basis for evaluating test completeness.
In \cite{DecMaking}  resp. \cite{SituationAssessment} uncertainties of sensor data are modelled in a Bayesian approach for situation assessments and decision making. In  \cite{huang_ontology-based_2019} this is extended with an ontology that capture different aspects of the driving scene, such as the objects, events, and context. The ontology contains logical rules and constraints that record the knowledge and experience of human drivers and experts. However probabilistic dependencies among the various entities, conditions, and features are not captured in their models. Their selections remain uncorrelated, only logical interdependencies are captured. Our approach extends this approach with probabilistic modeling, which also allows us to make statistical observations.

A more recent paper \cite{Peleska.21.12.2023}  also employs statistical methods to assess the quality of neural networks. 
On the one hand, a model-agnostic approach is adopted. In a manner analogous to the refined statistical model proposed in  chapter \ref{section:StatistObs} they use a Bernoulli experiment to statistically model the test and obtain coarse estimates of sample sizes.  
In contrast to our approach, they assume complete and representative training and validation data, which precludes the possibility of detecting potential imbalances. In another approach, they take a white box view of the system under test (SUT), in their case a convolutional classification net.
The authors posit that they are able to partition the input space, defined as the equivalence classes of pre-images, with respect to the classification produced by the net. This is achieved by extending the point set of classified pre-images by utilising locally 1-connectivity of the set due to a.e. differentiability of the net. Based on this partitioning, the authors obtain stronger estimates of the sample sizes. The effectiveness of this approach is demonstrated in the case of MNIST.
It is uncertain whether sufficient labelled training data are available for complex input spaces and classifications. Furthermore, for complex image sets, the topologically 1-connected subsets of preimages with the same classification may not form a feasible and manageable partitioning. We would like to express our gratitude to Jan Peleska for providing us with this work.

In their approach \textsc{PROMPTATTACK}, the authors \cite{Metzen.09.03.2023} propose a different, AI-supported method for identifying semantic subgroups of input data that are particularly susceptible to misclassification. 
In contrast to the approach proposed here, they also assume given training data and data distributions and therefore cannot reveal weaknesses hidden in these assumptions.
Conversely, the derived methods for detecting systematic misjudgements of ML systems can be employed in conjunction with our approach.

\subsection{Structure of this work}
\label{section:Structure}
Conventional software systems operate deductively, in accordance with established rules and the specifications imposed upon them. They are implemented imperatively. In contrast, ML-based systems learn inductively, based on a multitude of examples. They function statistically, in accordance with the principles of probability theory.
Chapter \ref{section:ConvVsML} elucidates the distinctions between the two systems and presents conclusions regarding testing.

In general, successful tests give us confidence in the quality of a system. Therefore, they should cover the most important aspects of the system and its application domain.
In the context of machine learning (ML) systems, ontologies are frequently developed with the objective of enhancing their completeness, see see \cite{ASAMOpenSCENARIO,scenic,dissSchuldt,CoreOnt2015,Breitenstein.2020,Bogdoll.01.09.2022,Breitenstein.11.02.2021,Guneshka22,Zipfl.21.04.2023,UlHaq.2019}.
Our approach is also informed by this perspective, chapter \ref{section:Ontology}. 
But simple experiments show that these approaches are not sufficient, see chapter \ref{section:N-wise-testing}.

In light of these considerations, we propose an extension of this approach with a novel concept of \textit{Probabilistically Extended Ontology}(PEON).
This is discussed in greater detail in chapter \ref{section:ProbOntology}.Furthermore, we demonstrate that the identified issues can be effectively addressed with the aid of this framework. This is presented in chapter \ref{section:ExpResults}.

In chapter \ref{section:StatistObs} we derive a statistical model for the testing process from this new concept.
This enables us to specify the sample size and variability of test cases and define end-of-test criteria based on the quality requirements for the system.

In the chapter \ref{section:TestGen} we demonstrate how this novel concept can be effectively integrated into an automated test chain for an object recognition system. Such a test pipeline enables block box validation of an ML-based system that is independent of development and training. 

We conclude this paper with an outlook on future planned work.

\section{Testing ML-based systems}\label{section:ConvVsML}
For traditional software systems, there are usually sufficiently detailed requirements and definable areas of application, whereas for most ML-based systems, we are dealing with open, barely comprehensible targeted environments and areas of application.
As an example, consider possible road scenes for object detection, where it is no longer possible to specify sufficiently comprehensive rules as to how the system should behave.
The traditional software systems that operate deductively, in accordance with predefined rules, and are implemented imperatively are unable to fulfil the requisite functions and are consequently replaced by ML-based systems that learn inductively from a multitude of examples. These systems function statistically, in accordance with the principles of probability theory, see table \ref{tab:ConvVsML}.

\begin{table*}[h] 
   \centering
   \begin{tabular}{@{}lcl@{}}
     \toprule
     Conventional embedded systems & \hspace{0.2cm}    & ML-based systems \\
     \midrule
     Deterministic System (Controlled Input/Output Relation)&  & Statistical System, Statistic Inference    \\
     Implemented Deductively, Rule-based && Inductive Learning (based on Training Data)\\
     Imperative Coding (C, Matlab,\dots) &  & Configuring and Hyperparameterizing Neural Nets,\\ 
     &  & Adaptive Controls  
     \\
     Controlled Bug Fix possible (every Bug found in Test 
     &  & Bug Fix barely possible, Relearning 
     \\
       should be fixed)&  &
     \\
     Extensive, Detailed Requirements
 &  & Poor Requirements (open ODD)
 \\
         \bottomrule
   \end{tabular}
   \caption{Conventional Embedded Systems versus ML-based Systems}
   \label{tab:ConvVsML}
 \end{table*}
Consequently, the behaviour of an ML-based system is based on statistical inference, which is derived from Bayesian rules, MAP or maximum likelihood\dots  \cite{Goodfellow-et-al-2016}. 
It is crucial to acknowledge that statistical errors are inherent to statistical analysis and that specific instances of misconduct are challenging to rectify, as there are no explicit rules or algorithms that can be corrected in this context. 

In traditional software engineering, testing is conducted independently of development. 
Practitioners have access to a range of established methods that enable them to assess the extent to which a test approach is comprehensive and exhaustive. 
They can fall back on a variety of established methods that allow them to measure the coverage and completeness of a test approach, this is not yet the case with ML. 
Before examining the implications of the statistical nature for testing ML-based systems, it is beneficial to gain an understanding of how other complex systems are systematically tested. We will adopt some of their processes.

\subsection{Systematic testing of conventional systems}\label{section:SysTestConvES}

Embedded systems are typically characterised by a large number of input parameters, the combinations of which are almost impossible to test. They are highly complex.
\Ex
The efficacy of an automotive braking system is to be evaluated. The current speed and the road and visibility conditions are among the key input parameters. The sheer number of potential input parameters is vast, and a systematic approach is necessary to ensure the effectiveness of the testing process.
\Exa

In order to test such a system, we divide the possible domains for the input parameters into disjoint sub-ranges (partitioning testing). In this case, the question arises how differentiated the areas should be. In the context of systematic testing of embedded systems or the equivalence class method, this question is answered by the Uniformity Hypothesis.

\noindent\textbf{Uniformity Hypothesis:}
\noindent\textit{The test result is independent of specific selection of value from value assignment range}

\Ex
In the case of the braking system and the input parameter \textit{speed} with the ranges \textit{[0,5]km/h, [5,30]km/h, [30,80]km/h, [80,200]km/h}, this hypothesis postulates that the test outcome will not be influenced by the selection of speed to 3.1 km/h or 4.9 km/h, both of which fall within the range '[0,5] km/h', or 35 km/h or 72 km/h, both within the range '[30,80] km/h'. 
\Exa 

More precisely, the hypothesis states that two test cases that have the same combination of features (parameters drawn from specified sub-ranges) should exhibit the same behaviour (hence the designation: Equivalent Behaviour of Test Cases $\sim$ Parameters in Equivalence Classes for the Test Cases).
We will see in the chapter \ref{section:SystematicTest}, that we can postulate a similar hypothesis in the context of the systematic testing of ML-based systems.

One of the most popular methods in the automotive domain utilising this approach is the classification tree method, which was developed and extended to systematic testing in \cite{GrimmGrocht}. A tool to support this method, \cite{CTE}, has been implemented.
In essence, each input parameter is associated with a node in a tree. 
Its children are then disjoint input ranges of its possible values. For the abstract test, each input parameter is then assigned exactly one value range from which the concrete values are selected. 
This methodological approach has been demonstrated to be both successful and effective in the testing of embedded software, and has already been applied on numerous occasions. 

In view of the high complexity and variability of the input data, ontologies are used to describe ML-based systems instead of classification trees.

\subsection{Ontologies}\label{section:Ontology}
In contrast to tree structures, taxonomies, and data models, ontologies are capable of storing and checking the consistency of semantic knowledge, as well as deriving new facts from explicitly specified facts, even in cases of considerable complexity. 
Their use is envisaged in future standards, \cite{EASA.01.11.2023}, for example to provide a reference to the completeness of training data.

Ontologies are developed to describe specific application domains of ML-based systems, the Operational Design Domain (ODD), see \cite{ASAMOpenSCENARIO,scenic,dissSchuldt,CoreOnt2015,Breitenstein.2020,Bogdoll.01.09.2022,Breitenstein.11.02.2021,Guneshka22,Zipfl.21.04.2023,UlHaq.2019}. 
In essence, ontologies describe classes (potential entities in an ODD) and their relationships, as well as their possible characteristics and instantiations (properties of the entities). 
Fundamentally, ontologies contain the essential concepts that we are interested in.
If we are interested in detecting obstacles in railway yards, the ontology will differ from the one we would develop for detecting bad weather factors and their influence. This is consistent with the fact that any machine learning (ML) based system must be trained on the ODD.

Figure \ref{fig:ProbOntology} depicts a highly simplified portion of an ontology, which is intended to illustrate the possible individuals present in a street scene. 

The classes are \textit{sex, age, height} and their possible characteristics are roughly divided into ranges like \textit{>1.8 m, [1.65, 1.8]m, [0.8,1.65]m, < 0.8 m}.

\begin{figure}
   \centering
   \includegraphics[width=0.3\textwidth]{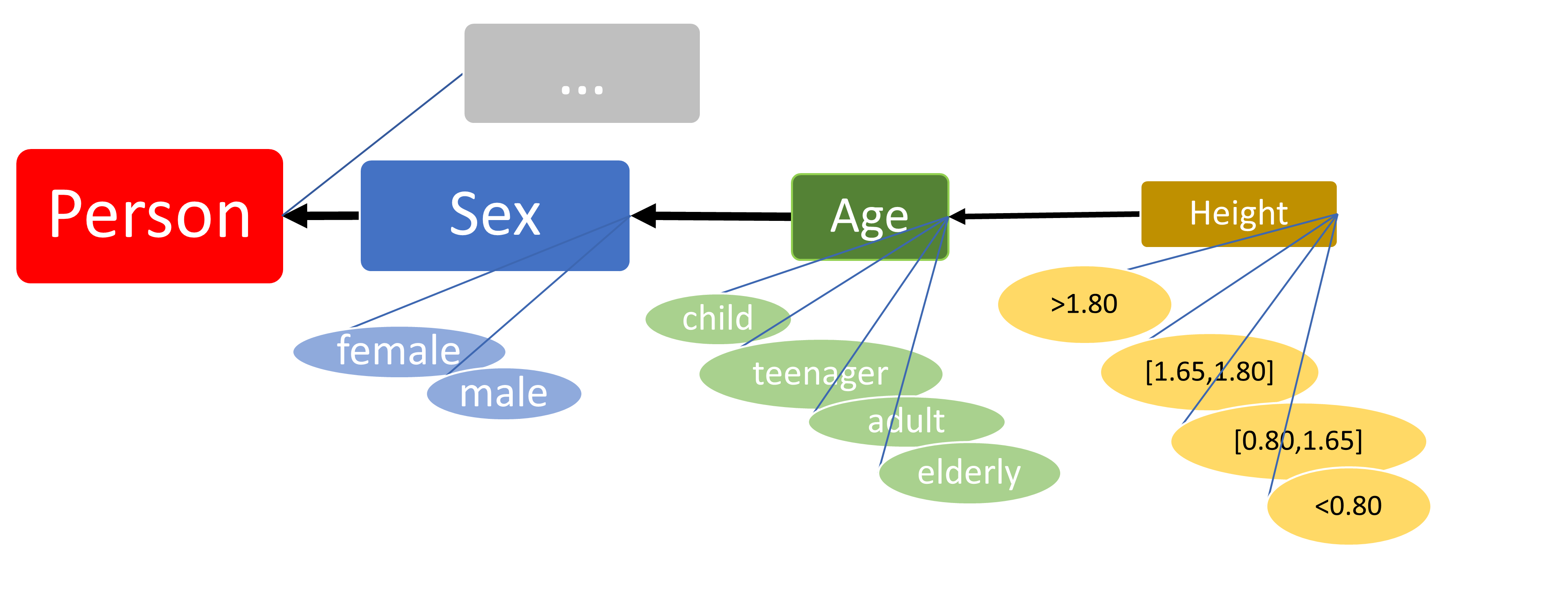}
   \caption{Simplified ontology specifiying a person} \label{fig:Ontology}
\end{figure}

Such an investigation also allows for the systematic identification of edge cases that require consideration. The completeness of the training data can then be defined as the requirement that every possible concept permitted by an ontology is represented in the training data set. 
Nevertheless, the question of the optimal level of ontology sophistication and the necessity for further refinement remains unresolved. 

\Ex
Rather than dividing the height of an individual into four possible distinct domains, \{>1.8 m, [1.65, 1.8]m, [0.8,1.65]m, < 0.8 m\}, we could also use 5, 6 or 10 subdomains by splitting $[1.65,1.8]m \rightarrow [1.65,1.72]m \vee [1.72,1.80]m $ etc.. What constitutes an appropriate refinement?
\Exa

It is evident that an ontology comprising an ODD defines a partition of the training data.
If, for each possible characteristic of a class, we select an associated property range from the ontology, for example, as age is a teenager, height is between 0.80 and 1.65 metres, and age is 15 years and height is 1.60 metres, we can group the data according to this combination.
The set of all possible combinations permitted by the ontology thus encompasses all potential data (completeness) and divides it into mutually exclusive groups. These groups are designated as \textit{partitions} analogous to the systematic test in conventional software testing, see \ref{section:SysTestConvES}.

\begin{figure}
   \centering
   \includegraphics[width=0.4\textwidth]{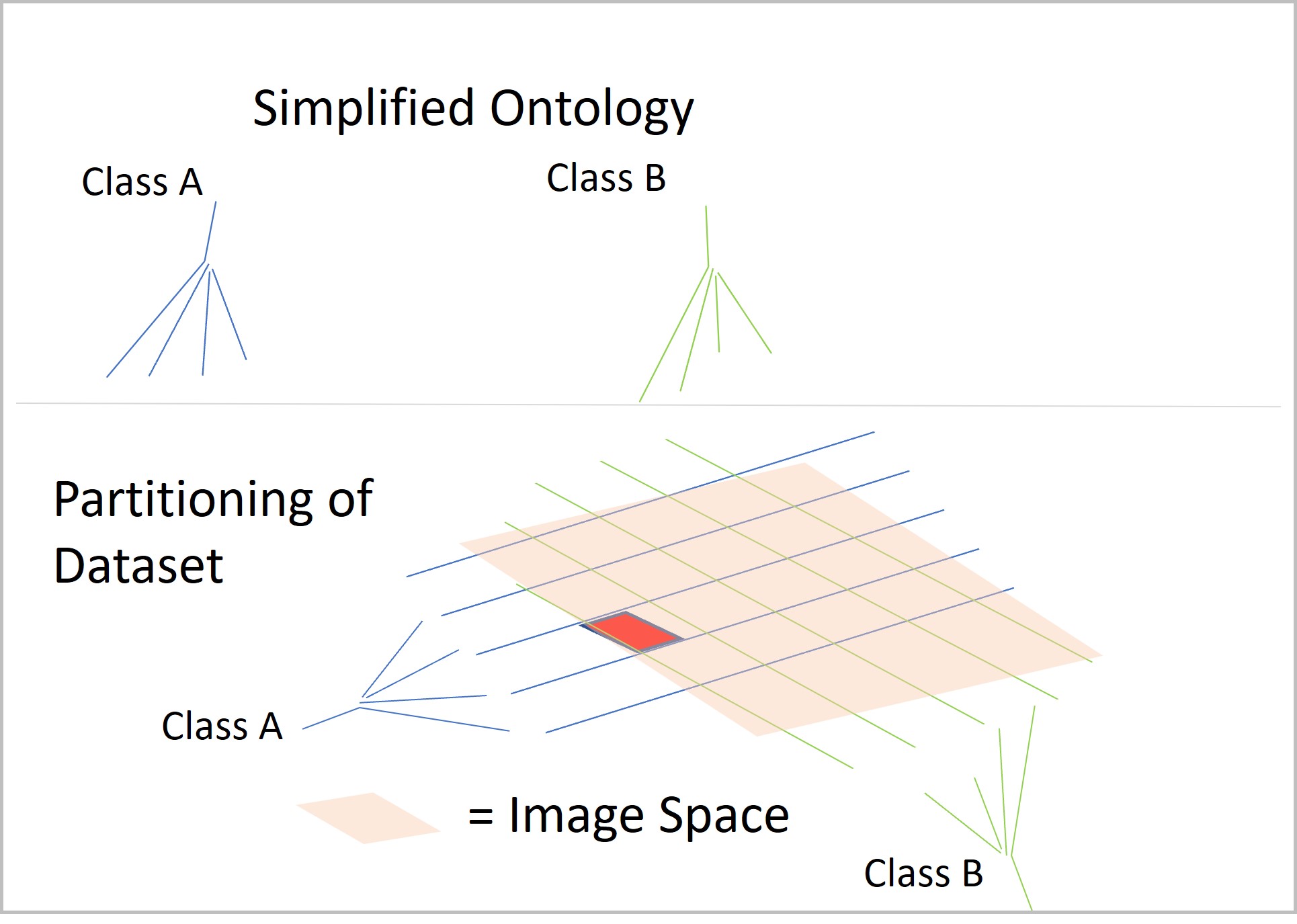}
   \caption{Partitioning of a dataset according to an ontology} \label{fig:OntologyPartitioning}
\end{figure}

In this paper, we will employ the terminology of ontologies. 
In this context, the term \textit{classes} denotes the possible entities corresponding to input variables while the term \textit{instantiations} or \textit{properties} denotes their possible characteristics or value ranges as is illustrated in the context of embedded systems.

\subsection{Statistical nature of ML-based systems}\label{section:MLbasedTesting}
However, partitioning methods such as classification trees or ontologies do not take into account the inherently statistical nature of ML-based systems.

In the majority of cases, testing and validation are considered to be part of the training process. The training data set is divided into two distinct subsets: training data and validation data. The latter is not used for training purposes but is employed to assess the quality of the trained system, see \cite{Geron.2019d}. 
The specified quality criteria are typically accuracy, precision, statistical averages of misclassifications that have occurred.

In accordance with the statistical nature of ML-based systems, the quality criteria are statistical in nature too: 
\be
accuracy = \frac{\sum_{\mbox{all test cases}} (\mbox{System Response } == \mbox{correct})}{N} 
\ee
with $N=$ total number of test cases.

Misbehavior resulting from unbalanced or insufficient training data cannot be detected by the final quality assessment with this approach. For this reason, standards today \cite{EASA.01.11.2023}, \textit{Chap. 3.1.3.8. Data validation and verification }.
The necessity for an independent approach is becoming increasingly apparent.

The evaluation criteria are interpreted as statistical statements about the safe operation of the system in its targeted environment (ODD),for example, the reliability with which a person on the street will be recognised by the system.
In order for such a conclusion to be permissible, the set of test cases must fulfil  a number of properties, as outlined in list list \ref{it:tcprops}.
These properties are of particular importance.
\begin{itemize}\label{it:tcprops}
  \item \textit{complete}, i.e. cover all possible street images,
  \item \textit{representative, balanced}, i.e. reflect the distribution of real street scenes well,
  \item extensive enough to allow valid statistical statements to be made.
\end{itemize}

\Nb
The potential risks associated with incomplete or imbalanced training data are well documented, as evidenced by the findings presented in \cite{EffectSepSampling}.
It is therefore evident that the training data should also exhibit these characteristics.
Although this paper primarily focuses on the testing phase, our approach can also be employed to enhance the training data, see \ref{note:TraingData}.
\Ne
\subsection{Statistic of N-wise Testing} \label{section:N-wise-testing}
A common approach to testing conventional systems is combinatorial testing. This involves attempting to cover a large number of different input combinations with a minimum number of test cases. A test data set is said to fulfil the pair-wise condition if every combination of two input parameters is present in the data set at least once. In general, the set fulfils n-wise if any combination of n parameters is present.
Such a set of test cases is complete with respect to the partitioning.

We can then calculate pass/failured test case statistics over a combinatorial selection of test cases.
In this chapter, we will demonstrate the potential errors that can arise in this process through a simple experiment.

The system under investigation employs an ML-based object detection system, based on the Coco datasetbased on the Coco dataset (\cite{linMicrosoftCOCO2015}), in conjunction with the centernet hourglass model.

\begin{figure}
  \centering
  \includegraphics[width=0.5\textwidth]{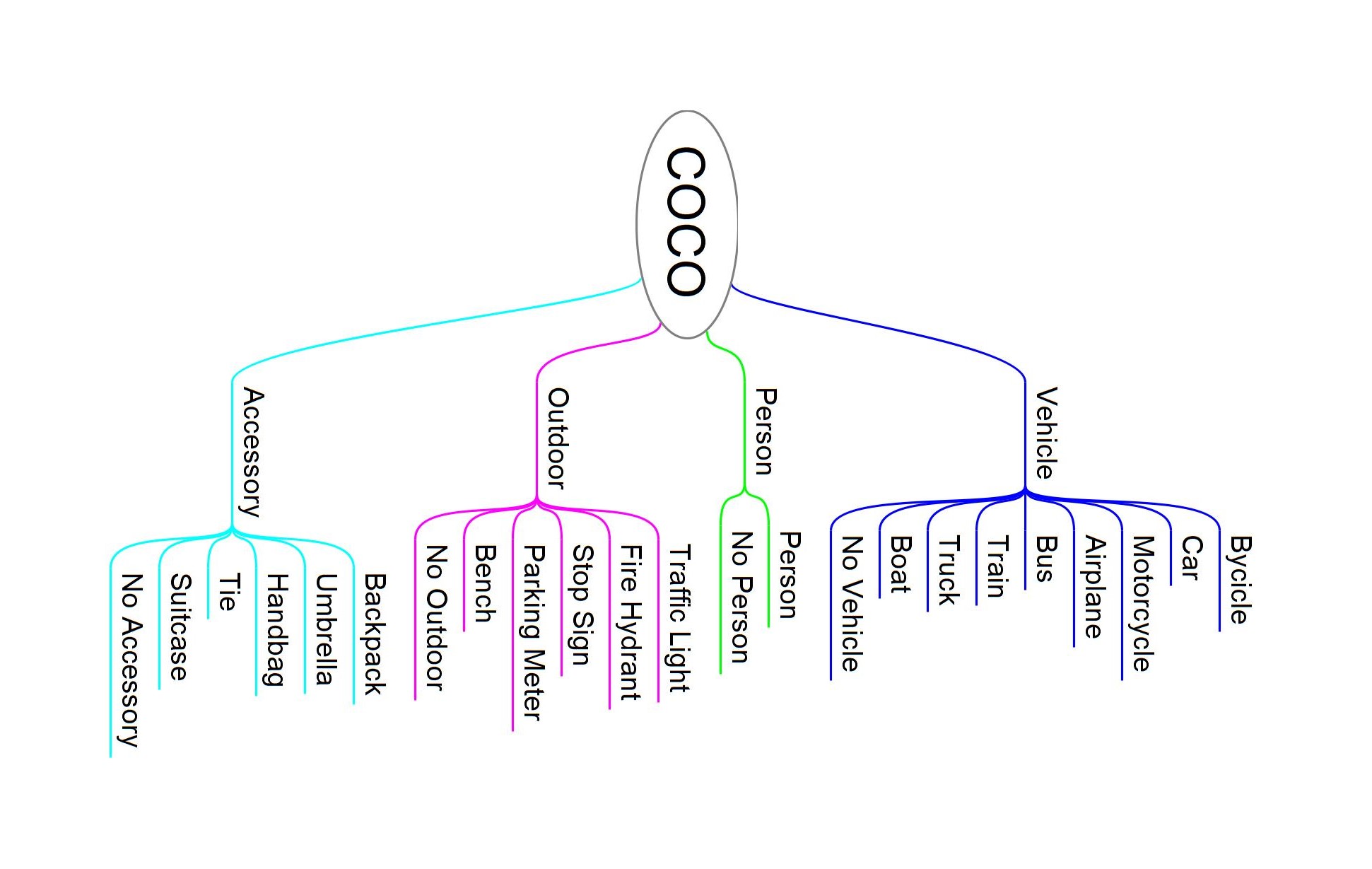}
  \caption{COCO ontology} \label{fig:COCOOntology}
\end{figure}

A rudimentary ontology was devised, comprising the following classes: \textit{person, vehicle, outdoor, accessory}, figure \ref{fig:COCOOntology} depicts this ontology

\begin{figure}
  \centering
   \includegraphics[width=0.4\textwidth]{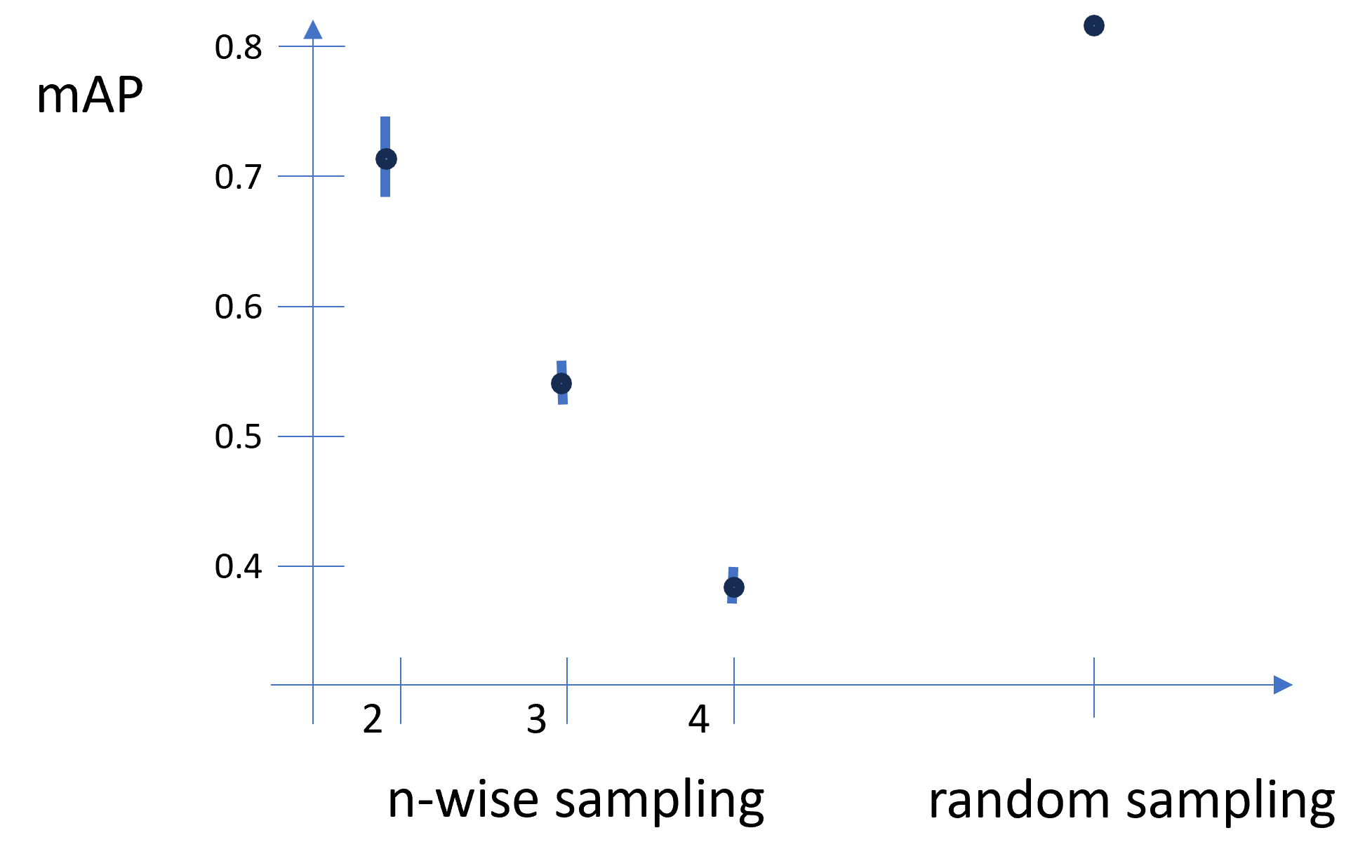}
   \caption{2-, 3-, 4- wise Combinatorial Testing vs. random sampling} \label{fig:ExpResults}
\end{figure}

This ontology has been employed to select training images. 
The input variables are configured according to the classes of our ontology and their assignment properties. 
In other words, given a combination of values for classes (properties such as Age $\rightarrow$ children), we select a training image that satisfies this labelling.  Finally, we tested the quality of the system using these test data, see figure \ref{fig:ExpResults}
(In order to ensure the reliability of the data, we consistently sampled the same number of test data and repeated the test series randomly.).

The selection of test data is not representative, with an unbalanced distribution of selections. 
The distribution of the test data follows more or less the uniform distribution along the ontology. 
As the restrictions become stronger, from two-wise to three-wise to four-wise, the test data becomes less reflective of the training data.
In particular, the results of this experiment demonstrate that N-wise testing will never reproduce the key criteria for quality in the case of ML-based systems as given to us by the developer.
Such an approach is not consistent with development.

\section{Probabilistic extension of ontologies}\label{section:ProbOntology}

The experiment presented in chapter \ref{section:N-wise-testing} demonstrates that a purely combinatorial approach to the static evaluation of the test of ML-based systems is not consistent and insufficient.
And as previously discussed in chapter \ref{section:ConvVsML}, it is of paramount importance that the set of test cases is not only exhaustive but also representative, balanced, and comprehensive, see the list \ref{it:tcprops}.
The concept of representativeness pertains to a ground set upon which a statistical statement is to be made. 
In the case of ML-based systems, this is typically an ODD, which is comprised by an ontology. 
The representativeness or balance then refers to a fictitious distribution of possible situations within the ODD.

\begin{figure}
   \centering
   \includegraphics[width=0.4\textwidth]{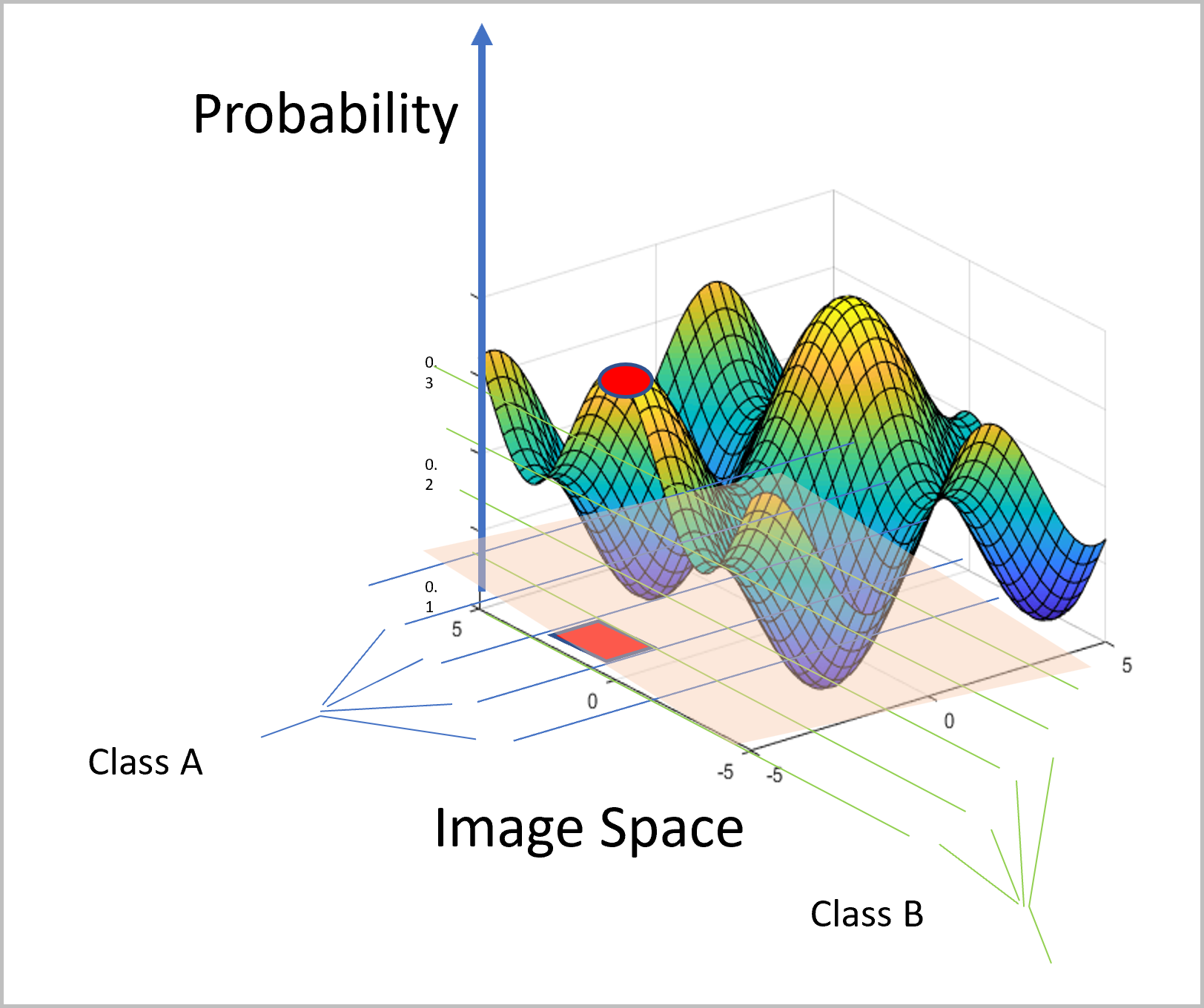}
   \caption{Probabilistic distribution over some dataset} \label{fig:ProbOntology}
\end{figure}

We propose an ontology that describes the ODD. As outlined in chapter \ref{section:Ontology}, this naturally leads to a partitioning of the input space of our ML system.
A discretisation of the space is obtained, as illustrated in figure \ref{fig:OntologyPartitioning}.
Figure \ref{fig:ProbOntology} shows then the distribution of the data approximated as a probability distribution over the partition space.
The representativeness or balance of the test data can be described informally as follows.
The sum of the probabilities of all data present in a partition P (\textit{prob(P)}) describes the frequency of a situation occurring in P.
A data set is representative if it contains data from P proportional to \textit{prob(P)}.
In other words, consider the partition marked in red in the figure \ref{fig:ProbOntology}. 
This partition represents a data set with as many examples as the function value marked in red. This is true for all partitions.

Our approach can now be carried out from an ontology. The derived partitioning provides a discrete model of the ODD.
Let $X^i, i\in{0,1,\dots n}, n\in \N$ be the set of all classes, potential entities in the ODD.
Let $x_j^i, i\in{0,1,\dots n}, j\in {0,\dots,n^i}$ denote the corresponding subsets of properties of $X^i$,  according to the ontology, i.e.
$X_i = \{x_0^i,x_1^i,\dots x_{n^i}^i\}$.
The product space, $\Pi_{i=0}^{n} X_i$, defined as the Cartesian product of the  $X_i$, covers the basis space of all partitions.
In this notation, a partition P is represented by a tuple of the form $(x_{j_0}^0,\dots x_{j_n}^n)$.

A probability distribution on the space of all partitions is then a mapping

\equ
p: \Pi_{i=0}^{n} X_i &\rightarrow & [0,1] \hspace{0.1cm}\mbox{probability distribution} \\
(x_{j_0}^0,\dots x_{j_n}^n)& \mapsto & p(x_{j_0}^0,\dots x_{j_n}^n) , j_k\in \{0,\dots n^k\} \\
\eque

with
\be
\Sigma_{i\in\{0,\dots n\}} \Sigma_{j_i \in\{0,\dots n^i\}} p(x_{j_0}^0,\dots x_{j_n}^n) =1 \hspace{0.1cm}\mbox{normalization}
\ee

For a partition P, the  probability is then given by the following expression \textit{prob(P)} $= p(x_{j_0}^0,\dots x_{j_n}^n)$

A \textit{probabilistically extended ontology} is now defined as an ontology together with a probability distribution p on its derived partitioning as described above.
This extension allows the uniformity hypothesis of the systematic test of conventional systems, as described in chapter \ref{section:SysTestConvES}, to be adapted to the following postulate.

\noindent\textbf{Uniformity hypothesis:}
\noindent\textit{The probability of a negative test result is independent of the specific instantiations of the classes when the properties are selected from the same partition.}

In other words, the probability of a test associated with a partition failing is the same for all tests associated with that partition. 
This criterion allows us to determine how much the ontology should be refined. 
It means that we should refine the partitioning until we expect a constant probability of failure for any representative therein. 

This postulate allows us to construct a statistical model for the test. 
Let $p_P\geq 0$ be the probability of failure for a test lying in P.
Then the execution of the test can be described statistically by a Bernoulli experiment with parameter $p_P$, see \cite{Dudley.2002}.
There is a probability $p_P$ that the test will fail.
The opposite is the probability $\varphi_P$ that a test will be in partition P.
The total statistical model of the test is then described as the sum of Bernoulli experiments 
\[
\mbox{statistical model of testing } = \sum_{i=0}^{n} \varphi_{P_i}(\cdot)\cdot\mathcal{B}(p_{P_i})
\].
A \textit{probabilistically extended ontology} thus partitions our ODD and allows a discrete stochastic model for testing.

\subsection{A constructive approach to probabilisticly extend ontologies}\label{section:ConstrProbExtension}
It is evident that the extension of a complex ontology probabilistically is a highly non-trivial endeavour. 
In order to address this, we developed methods and several modelling techniques to specify a probability distribution over complex ontologies and demonstrated these techniques for complex ODDs. 
In \cite{ProbExtOnt} we demonstrate, how this can be achieved, even in complex ontologies such as OpenXOntology 1.0.0,\cite{ASAMOpenXOntology}.
This chapter will present a preliminary sketch of the subject matter.
However, due to the limited space available in this article, we are unable to provide a more detailed account of this topic. Therefore, we refer you to the following source: \cite{ProbExtOnt}.

The initial step is to define the probability distributions over classes like \textit{Age} or the \textit{Height}.
These distributions are known as marginal distributions of the probability space.
In order to achieve this, one may consult statistical tables available online or utilise the central limit theorem to construct a model.
For example the \textit{height} of an individual is well modelled by a normal distrubition.

Modelling probabilistic dependencies between classes is a more challenging task. 
For instance, an individual's  \textit{height} is also dependent on their \textit{age}. 
In the paper \cite{ProbExtOnt} , we demonstrate how to specify marginal distributions for entities and their properties, as well as model Boolean or functional dependencies between them.

Having defined a \textit{Probabilistically Extended Ontologies}, a set of representatives for the partitions is sampled accordingly.
They build a set of abstract test cases.
Subsequently they are further enriched into concrete, executable test cases.

Nevertheless, developing a \textit{Probabilistically Extended Ontologies} is a complex and time-consuming task. 
However, it provides a transparent representation of the ODD, not only in terms of its potential entities, but also how the different realisations are distributed. 
This enables the operationalisation of what should be representative and balanced. 
In particular, we obtain an explicit probabilistic approximation of the target environment, the ODD. 

\Nb
The use of \textit{Probabilistically Extended Ontologies} (PEON) enables the assessment of representativeness and balance for the training data set. 
This is achieved by sampling from the aforementioned set. 
In the majority of cases, the labelling of the training data is not as finely structured as the partitioning of the ODD. 
In that case the data will be classified manually according to the ontology partitioning.
Subsequently, standard statistical tests (hypothesis and correlation tests, t-test, Chi-squared test, Mann-Whitney-U test, etc., see  \cite{Bijma.2017}) can be employed to substantiate the balance of the training data with respect to a selected subset of partitions. 

An analysis of the Bernoulli coefficients $\overline{p}_h$ of partitions also allows us to detect weaknesses in our ML-based system.

\Ne

In a future work we will describe these ideas in more detail.

\subsection{Some experimental results}\label{section:ExpResults}
In the majority of cases, the focus of testing is limited. 
This is exemplified by the need to ascertain that the system continues to function effectively in adverse weather conditions or that it can only be used on the motorway. 
This fits in with an iterative approach to a step-by-step refinement of an ontology and probabilistic extension.

We conducted another simple experiment, see \cite{OnSysTestExp}.
A classification network was trained with the PEdesTrian Attribute dataset (PETA), \cite{PETA}.
An ontology was designed for the labels of this dataset and extended probabilistically in two ways. First, the distribution of individual labels (marginal distributions) for the training data was computed. Second, marginal distributions for individual labels were modeled along with various conditional expectations between labels, based on statistical data from the web and the authors' own experience.

\begin{figure}
  \centering
  \includegraphics[width=0.6\columnwidth]{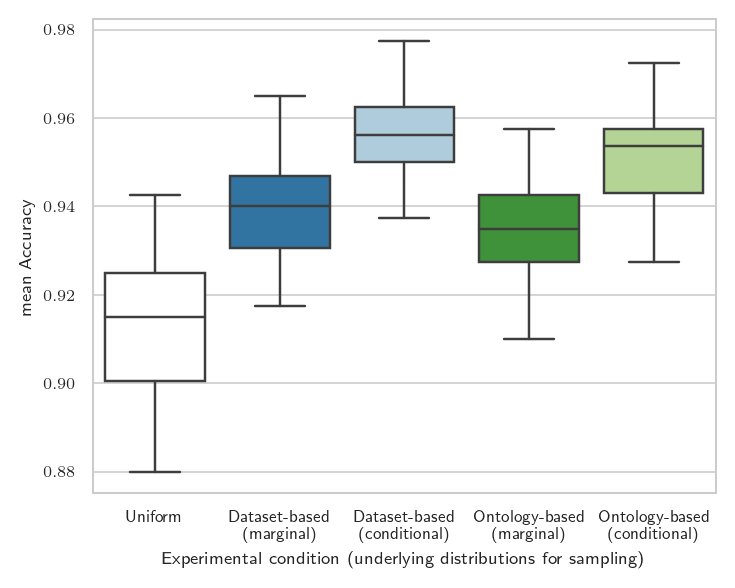}
  \caption{Mean average accuracy for different experimental conditions from the dataset or from the marginal/conditional distributions. The plot shows the overall accuracy.} \label{fig:samplingresults}
\end{figure}

We then selected test sets from the training data set in various ways:
\begin{itemize}
  \item uniform (combinatorial)
  \item Peta dataset based marginal distribution (only respecting the marginal distribution as given in the Peta set)
  \item Peta dataset based distribution (random from Peta)
  \item Ontology-based estimated marginal distribution (only respecting the marginal distribution as estimated)
  \item Ontology based, full estimated distribution (randomly according to the estimated probabilistically extended ontology)
\end{itemize}
and performed the classification tasks on the different test set selections.
Our results are summarised in Figure \ref{fig:samplingresults}.

The developer gives us an accuracy of approximately 0.955 (light blue). 
When this is taken as a reference, it can be seen that with a uniform selection (white), significantly different measurements are obtained. 
However, when the marginal distributions are taken into account, the results improve significantly. 
It is of little consequence whether the marginal distributions are taken from the data set (dark blue) or from the estimates (dark green). 
Furthermore, the results are even better when the various dependencies are also modelled (light green).
The experiment serves to illustrate the consistency of our approach with the statements of the developer in the case of identical data.
For more detailed information the reader is referred to \cite{OnSysTestExp}.

This straightforward experiment indicates that more accurate quality measurements can be achieved through improved modelling of probability distributions.
\subsection{End-of-test criteria for ML-based systems}\label{section:StatistObs}
The objective of testing ML-based systems is to ascertain the statistical quality criterion.
\textit{Probabilistically Extended Ontologies}
gives us a statistical model of the test, see chapter \ref{section:ProbOntology}. This model permits the estimation of the statistical significance of tests. Conversely, it can also be employed to calculate the number of tests that must be performed in order to achieve a specified quality level.

A random sample of test cases may contain a statistical bias with respect to the underlying distribution from which the sample was taken. 
In order to statistically mitigate the bias, it is necessary to have a sufficiently large sample size, which depends on the distribution.  
As explained in the chapter \ref{section:ConvVsML}, the goal of testing ML-based systems is to prove their quality statistically.
It is therefore essential to perform a sufficient number of tests to avoid Type I or Type II errors. These are instances where the quality claim is rejected even though the system meets it, or where a claimed quality is accepted even though the system does not meet it.
The relevant significance parameters are \textit{$\alpha$ } and \textit{power} for the statistical test, see \cite{Bijma.2017}.
In sloppy terms, if we repeat our testing process 100 times and the developer's quality claim is false, then the hypothesis should be rejected at least $(1-\alpha)*100 $\% of the time, \cite{Bijma.2017}. Conversely, if the quality claim is true, then it should be confirmed in (power*100 \%) of the cases.
It is important to note that the statistical quality of our tests, indicated by the parameters \textit{$\alpha$ } and \textit{power}, should not be confused with the results of the tests themselves.
In general, the sample size is dependent on the probability distribution and the required significance level, which is determined by the parameters  \textit{$\alpha$ } and \textit{power}.

If we restrict our analysis to a single partition, the distribution of test results can be described as a Bernoulli experiment $\mathcal{B}(p_{P})$, as discussed in chapter \ref{section:ProbOntology}.
The developer has asserted that the probability of failure is within the range $p_0 \% \pm 0.1*p_0 \%$.
Our objective is to test the veracity of this claim.
We set the significance to standard parameters $\alpha =0.2$  and $power = 0.8$. 
Once these values have been established, we can then look up the required sample sizes as a function of the Bernoulli parameter $p_0$, see \cite{Bijma.2017}.
This gives the minimum number of experiments required to achieve the desired level of confidence.

For example, if the probability of success, $p_0$, is assumed to be 0.05 with a standard deviation of 0.005 and significance parameters $\alpha$ =0.2, \textit{power} = 0.8, then a minimum of 9326 tests are required, with a maximum of 513 failures permitted.
This gives us, in the simplest case, a well defined end of test criterion.

Extending this result to the entire ODD would result in an excessively large number of test cases. 
However, this is not necessary. On the one hand, test data generation can be employed to enhance the number of test cases, as will be demonstrated in the subsequent chapter \ref{section:TestGen}. 
On the other hand, the utilisation of more sophisticated estimators is recommended. 
On the other hand, we should use more sophisticated estimators.

Our statistical model for testing is based on the \textit{Uniformity Hypothesis} and \textit{Probabilistically Extended Ontology},  which is a multinomial Bernoulli experiment, chapter \ref{section:ProbOntology}.
This structure suggests the use of stratified random sampling along the partitions, see \cite{Sarndal.2003}.
Estimators for the mean (accuracy) and its variance are obtained by
\bea
\mu &\sim& \sum_i \varphi_{P_i} \cdot \mu_i  \\  
var &\sim& \sum_i \varphi_{P_i} \cdot \mu_i\cdot(1-\mu_i)
\ea
where the mean failure rate on $P_i$, denoted by $\mu_i$, is an unbiased estimator for $p_P$, see \cite{Bijma.2017}.
In practice we estimate
\be \label{equ:Random}
\mu_i \approx max(\frac{\mbox{failed test case in }P_i}{\mbox{test cases in }P_i},
\frac{1}{\mbox{test cases in }P_i +1}).
\ee
The optimal allocation of sampling (Neyman allocation), namely the optimal number of test cases to be selected in $P_i$, is then estimated
\noindent 
\be \label{equ:Neyman}
n_i \sim  \frac{\varphi_{P_i} \cdot \mu_i\cdot(1-\mu_i)}{\mu\cdot(1-\mu)}
\ee
\noindent where $\mu$ is the accuracy as given by the developer.
For further details, please see \cite{Sarndal.2003}.

Firstly, it is necessary to carry out a random selection of tests in order to estimate the value of the Bernoulli parameter $p_{P_i}$ as accurately as possible, with reasonable effort, see equation \ref{equ:Random}.
The number of test cases required to evaluate the ML System can then be estimated using equation \ref{equ:Neyman}, which provides a solid basis for a statistical evaluation of the tests.
In this way we get a well defined and reliable end-of-test criterion.
We will come back to this point in future work, see \cite{WiesbrockEtAl}.

\subsection{Ethical points of view}
As described in chapter \ref{section:Ontology}, an ontology always reflects our interests.
To illustrate, if we demand, for ethical reasons, that skin colour should not play a role, this is reformulated as a demand for a uniform distribution on the space of skin colours. 
The same is true for age and sex. 
This means that the distribution underlying our test case selection must be uniform with respect to these properties.
Conversely, a \textit{probabilistically extended ontology} can be used to assess the consistency of the ontology with ethical principles.

\section{A Probabilictically extended ontology and testdata generation }\label{section:TestGen}
\begin{figure}[htbp]
\centering
\includegraphics[width=0.5\textwidth]{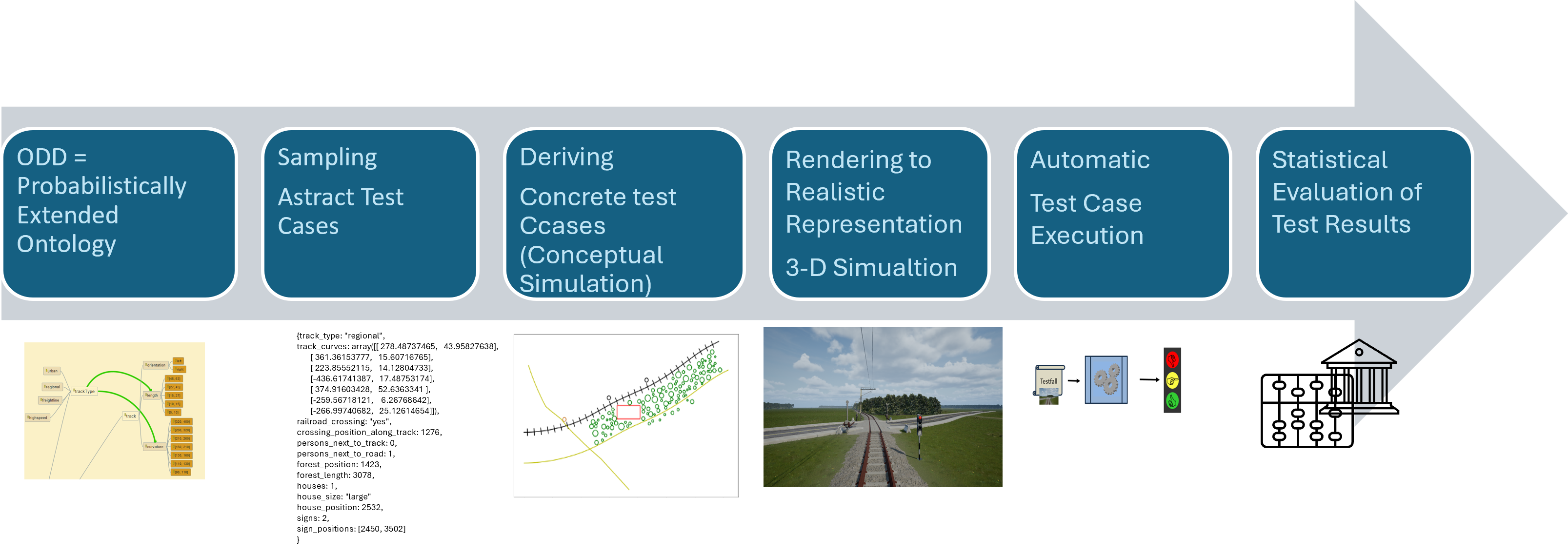}
\caption{Test chain with test data generator based on a PERSON}
\label{fig:TDGen}
\end{figure}

In a joint research project entitled \textit{KI-LOK} (Fraunhofer Fokus, Neurocat, Heinrich-Heine University Düsseldorf), we are investigating the possibility of certifying an object recognition system in the railway sector.
For this purpose, we have set up a test pipeline in which test cases for detecting objects along different rail tracks can be generated and executed on a system under test (SUT).
The construction of test cases eliminates the need for complex annotations. Ground truth for recognizing and locating the objects is available.
In order to obtain valid statistical test evaluations, abstract test cases are generated from a \textit{Probabilictically Extended Ontology}, (PEON,) and enriched to form concrete test cases.
Subsequently, these are rendered in a three-dimensional simulation and presented to the system under test, see figure \ref{fig:TDGen}.

A demonstrator of our approach will be available at the beginning of autumn 2024.

\section{Conclusion and future work}\label{section:Conclusion}
This article is based on the hyphothesis that for ML-based systems failed test cases cannot be directly interpreted as a bug that needs to be corrected. ML-based systems work probabilistically and counterexamples are always to be expected statistically. 
The quality requirements for ML-based systems are statistical in nature, and consequently the tests of ML-based systems need to take this into account. To address this fact, we propose the new concept of \textit{Probabilistically Extended Ontologies}.
Ontologies assist in the capture of the concepts of complex environments.
The probabilistic extension enables the definition of statistical metrics that apply to rigorous black-box tests for ML-based systems, as well as the establishment of operational end-of-test criteria.

As previously stated, this concept facilitates the assembly of training data. Additionally, it provides a sampling procedure for obtaining related abstract test cases. In future work, we intend to extend the methods and tools supporting the development of \textit{Probabilistically Extended Ontologies}.
We will also demonstrate how these techniques can assist in the analysis of systematic weaknesses in such systems, in a manner analogous to \cite{Metzen.09.03.2023}. Furthermore, we will revisit the estimation of sample sizes for significant tests, with a view to providing more robust evidence.

\bibliographystyle{IEEEtran}
\bibliography{SysTest4AI.bib}

\end{document}